# Differential Search Algorithm-based Parametric Optimization of Fuzzy Generalized Eigenvalue Proximal Support Vector Machine

M. H. Marghny
Computer Science Department, Faculty of Computers and Information, Assiut University, Egypt.

Rasha M. Abd El-Aziz
Computer Science Department, Faculty of Science, Assiut University, Egypt.

Ahmed I. Taloba
Computer Science Department, Faculty of Computers and Information, Assiut University, Egypt

## ABSTRACT
Support Vector Machine (SVM) is an effective model for many classification problems. However, SVM needs the solution of a quadratic program which require specialized code. In addition, SVM has many parameters, which affects the performance of SVM classifier. Recently, the Generalized Eigenvalue Proximal SVM (GEPSVM) has been presented to solve the SVM complexity. In real world applications data may affected by error or noise, working with this data is a challenging problem. In this paper, an approach has been proposed to overcome this problem. This method is called DSA-GEPSVM. The main improvements are carried out based on the following: 1) a novel fuzzy values in the linear case. 2) A new Kernel function in the nonlinear case. 3) Differential Search Algorithm (DSA) is reformulated to find near optimal values of the GEPSVM parameters and its kernel parameters. The experimental results show that the proposed approach is able to find the suitable parameter values, and has higher classification accuracy compared with some other algorithms.

## 1. INTRODUCTION
Recently, information, growing in huge volumes creates the need to process large amounts of data. In order to find hidden patterns of data and convert them into useful knowledge, this is known Data Mining. This direction includes methods other than classical analysis, based on clustering analysis [1-4], classification analysis [5, 6], and solving problems of generalization, association and finding patterns [7-9]. This area of research has recently become more and more important.

Classification is the process of arranging data into homogenous group or classes according to some common characteristics present in the data. Support vector machine (SVM) has an excellent performance in many real life classification problems such as image processing, text classification and bioinformatics.

SVM which is an emerging data classification technique proposed by Vapnik in 1995 [10], and has been widely adopted in various fields of classification, nevertheless it suffers from complexity and parameters selection. A new method has been introduced in [11] by Olvi L. Mangasarian which called Proximal Support Vector Machine (PSVM). This method has solved the problem of complexity of standard SVM, but it suffers from poor performance in the case of noisy and unbalanced data. Recently an efficient approach to PSVM has been proposed also by Olvi L. Mangasarian which is called the Generalized Eigenvalue Proximal Support Vector Machine (GEPSVM) [12]. The complexity of standard support vector machine has been solved by GEPSVM. A fundamental difference between GEPSVM and SVM is that, GEPSVM solves two generalized eigenvalue problems to obtain two non-parallel hyper-planes, whereas, SVM solves one quadratic programming problem (QPP) to obtain one hyper-plane. Therefore, GEPSVM works faster than SVM. Experimental results in [12] showed the effectiveness of GEPSVM on some public datasets.

In real world applications data may affected by noise or error which significantly influences on the performance of GEPSVM. There are many approaches have been proposed by researchers for this problem [13-19]. More efforts are needed in order to improve the performance of the classification task in this type of data.

In addition, the major problems that are encounter in SVM and all its inferred methods are how to find near optimal values for the SVM parameters and select a SVM kernel as well as tuning its parameters. Unsuitable parameters setting lead to poor classification outcomes. Authors in [20-25] tried to find solution for SVM parameters. There are no particular method to find the optimal values for the SVM or GEPSVM parameters and kernel parameters. This problem is still an interesting topic for more research to find more appropriate values for GEPSVM parameters and kernel parameters.

For these reasons, an improved version of GEPSVM, called DSA-GEPSVM for short is proposed. A new method for computing the fuzzy membership function is used in the linear case. Furthermore, a new kernel is used in the nonlinear case. The new kernel is a combination between the polynomial and the radial base function kernel. For solving the problem of parameters selection, a new and powerful method which is called Differential Search Algorithm (DSA) [26-31] has been used. This makes the optimal separating hyper-planes obtainable in both linear and non-linear classification problems.

The remainder of this paper is organized as follows. In section 2 we briefly give description of the GEPSVM. Section 3 give description of the DSA. In section 4 the proposed method is described. Section 5 reports experimental results. Finally, the conclusions make up Section 6.

## 2. GENERALIZED EIGENVALUE PSVM
In 2006 Olvi L. Mangasarian and Edward W. Wild proposed the Generalized Eigenvalue Proximal Support Vector Machine GEPSVM [12] as a generalization of the SVM method. The new formulation does not need the planes to be parallel, but for each class, the algorithm finds a plane that is as close as possible to the points of one class and as far as possible to those in the other class. Due to the simplicity of GEPSVM, many researchers have refined it to improve the general performance of the classifier [32-35]. But GEPSVM still needs more improvements, and is a very good topic for researchers.

First we consider the classification problem of m points in the n dimensional real space $R^n$, represented by the $m_1 \times n$ matrix A belonging to class 1 and $m_2 \times n$ matrix B belonging to class 2, with $m_1 + m_2 = m$. For this problem, a standard linear SVM is given by a plane halfway between the two parallel bounding planes that bound two disjoint half spaces each containing points mostly of class 1 or 2 [12].

In MSPSVM the parallelism condition has been dropped, but requires that each plane be as close as possible to one of the data sets and as far as possible from the other one. Thus, we are seeking for the two planes in $R^n$:

$$P1: x'w^1 - \gamma^1 = 0,\ P2: x'w^2 - \gamma^2 = 0, \qquad (1)$$

where the plane *P1* is closest to the points of class 1 and furthest from the points in class 2, while the plane *P2* is closest to the points in class 2 and furthest from the points in class 1. Then the first plane of (1) is obtained by solving the following optimization problem:

$$\min_{(w,\gamma)\neq 0} \frac{\|Aw - e\gamma\|^2 / \left\|\begin{bmatrix} w \\ \gamma \end{bmatrix}\right\|^2}{\|Bw - e\gamma\|^2 / \left\|\begin{bmatrix} w \\ \gamma \end{bmatrix}\right\|^2}, \qquad (2)$$

where $\|.\|$ is the two-norm, and it has been assumed in [12] that $(w, \gamma) \neq 0$ which implies to $Bw - e\gamma \neq 0$. As introduced in [12] the numerator of the minimization problem (2) is the sum of squares of two-norm distances in the $(w,\gamma)$-space of points in the first class to the plane $x'w^1 - \gamma^1 = 0$, while the denominator of (2) is the sum of squares of two norm distances in the $(w,\gamma)$ space of points in the second class to the same plane. By simplifying (2) we can write,

$$\min_{(w,\gamma)\neq 0} \frac{\|Aw - e\gamma\|^2}{\|Bw - e\gamma\|^2}. \qquad (3)$$

Then Tikhonov regularization term is added [36] to reduce the norm of the problem variables $(w,\gamma)$ that determine the proximal planes (1). Thus, for a parameter $\delta$, problem (3) has been rewritten as follows:

$$\min_{(w,\gamma)\neq 0} \frac{\|Aw - e\gamma\|^2 + \delta\left\|\begin{bmatrix} w \\ \gamma \end{bmatrix}\right\|^2}{\|Bw - e\gamma\|^2}, \qquad (4)$$

We can rewrite (4) as follow:

$$\min_{z\neq 0} r(z) := \frac{z'Gz}{z'Hz}, \qquad (5)$$

where,

$$G := [A \quad -e]'[A \quad -e] + \delta I, \qquad (6)$$

$$H := [B \quad -e]'[B \quad -e], \qquad z := \begin{bmatrix} w \\ \gamma \end{bmatrix}. \qquad (7)$$

*G* and *H* are symmetric matrices in $R^{(n+1)\times(n+1)}$ and *I* is an identity matrix.

As pointed out in [37], the objective function of (5) is known as the Rayleigh quotient, hence its solution can be obtained by solving a generalized eigenvalues problem. That is, the eigenvector corresponding to the smallest eigenvalue can determine a plane effectively.

Similarly we can directly get the second plane by solving the following optimization problem.

$$\min_{z\neq 0} s(z) := \frac{z'Lz}{z'Mz}, \qquad (8)$$

where,

$$L := [B \quad -e]'[B \quad -e] + \delta I, \qquad (9)$$

$$M := [A \quad -e]'[A \quad -e]. \qquad (10)$$

L and M are again symmetric matrices in $R^{(n+1)\times(n+1)}$. As analyzed above, the two non-parallel planes can be obtained directly by solving the classical generalized eigenvalue problem.

The Nonlinear GEPSVM can be obtained easily by considering the problem of finding two non-parallel planes

$$P1: K(x', C')u^1 - \gamma^1 = 0,\ P2: K(x', C')u^2 - \gamma^2 = 0, \qquad (11)$$

where $C := \begin{bmatrix} A \\ B \end{bmatrix}$.

K is the kernel function, which will be presented in the next section. By employing the same regularization strategy as in equations (4-10), we can also obtain the two non-parallel planes by solving the optimization problems of the following equations:

$$\min_{z\neq 0} r(z) := \frac{z'Gz}{z'Hz}, \quad where\ z := \begin{bmatrix} u \\ \gamma \end{bmatrix}, \qquad (12)$$

$$\min_{z\neq 0} s(z) := \frac{z'Lz}{z'Mz}, \quad where\ z := \begin{bmatrix} u \\ \gamma \end{bmatrix}, \qquad (13)$$

where,

$$G := [K(A, C') \quad -e]'[K(A, C') \quad -e] + \delta I, \qquad (14)$$

$$H := [K(B, C') \quad -e]'[K(B, C') \quad -e], \qquad (15)$$

$$L := [K(B, C') \quad -e]'[K(B, C') \quad -e] + \delta I, \qquad (16)$$

$$M := [K(A, C') \quad -e]'[K(A, C') \quad -e]. \qquad (17)$$

G, H, L and M are symmetric matrices in $R^{(m+1)\times(m+1)}$.

Since 2006 GEPSVM has achieved great performance in many real live applications, but in some cases data may affected by noise and errors. Most classification methods give low classification accuracy with this kind of data, and need some modifications in order to increase the classification accuracy. One of the most effective ways to overcome this problem is by adding a fuzzy value to each training sample. The works of many researches carried out by adding fuzzy values to the standard SVM [13, 14, 17, 18]. Many attempts for adding fuzzy to GEPSVM have been illustrated as in [15, 16, 19]. A first attempt to obtain a fuzzy version of the GEPSVM classification is presented in [15, 16]. In [15] the authors attempt to solve the following problem:

$$\min_{(w,\gamma)\neq 0} \frac{\|S^A Aw - e\gamma\|^2}{\|S^B Bw - e\gamma\|^2}. \qquad (18)$$

With $S^A$ is the fuzzy membership weights for each point $A_i$ and $S^B$ is the fuzzy membership weights for each point $B_i$. $S^A$ and $S^B$ are diagonal matrices.

$$S_{ii}^A = 0.5 + \frac{e^{f(d(A_i, C_B) - d(A_i, C_A))/d_{AB}} - e^{-f}}{2(e^f - e^{-f})}, i = 1, \dots, p, \qquad (19)$$

$$S_{ii}^{B} = 0.5 + \frac{e^{f(d(B_i,C_A) - d(B_i,C_B))/d_{AB}} - e^{-f}}{2(e^{f} - e^{-f})}, i = 1, \dots, m. \quad (20)$$

Where $C_A$ and $C_B$ are the center of mass of the two classes, $d_{AB}$ is the distance between the two means, the function $d(.,.)$ is the Euclidean distance between two points, and f is a constant that determines the rate at which the fuzzy membership decreases towards 0.5. Another recently attempt of fuzzy GEPSVM can be found in [19] where the author proposed the following fuzzy function

$$S_{ii}^{A} = s + (1 - s).e^{-\left(\frac{\min(d(A_i,C_A)}{\max(d(A_i,C_B)}\right)^2}. \quad (21)$$

Where min $(d(A_i,C_A))$ is the minimum distance of the point $A_i$ from the centers in $C_A$, max $(d(A_i,C_B))$ is the maximum distance of the point $A_i$ from the centers $C_B$ of the other class, and s is a parameter weighting the contribution of the exponential term to $S^A$ ,for more detail see [19].

# 3. DIFFERENTIAL SEARCH ALGORITHM (DSA)

Differential Search Algorithm (DSA) is a recently and efficient evolutionary algorithm. DSA is effectively used to solve numerical optimization problems. The main idea of the DSA algorithm was inspired form the migration of superorganisms making use of brownian like motion [28].

Algorithms that make use of the principle of evolutionary computation are known as Evolutionary Algorithms (EA). These algorithms are suitable to search for the optimal (best) solution of many optimization problems. In real world problems the optimization process may have more than one solution, for searching for the optimal solution among all these solutions in a short time is a challenging task. If the search space is small then searching for the optimal solution will take short time. Working with data whose search space is very large is a challenge for most researchers. When the problem is very large with a great number of possible solutions, then finding the optimal solution is difficult. Evolutionary computation techniques are powerful and effective with this kind of data. EA includes the following techniques [28]:

- Ant colony algorithm
- Artificial Bee Colony (ABC) algorithm
- Cultural algorithms
- Differential evolution
- Evolutionary algorithms
- Evolutionary programming
- Evolution strategy
- Gene expression programming
- Genetic algorithm
- Genetic programming
- Harmony search
- Learnable Evolution Model
- Particle swarm optimization
- Self-organization such as self-organizing maps
- Swarm intelligence

The Differential Search Algorithm (DSA) is the most recent addition. There are a number of computational-intelligence algorithms that model the behaviors of the superorganisms [28-30]. In the present work DSA is used to get the best values of parameter values in the proposed algorithm, due to it has the ability to manage such problem. The pseudo-code indicating the function of DS algorithm is given in Appendix.

# 4. PROPOSED APPROACH

## 4.1 Linear Fuzzy DSA-GEPSVM

The proposed approach introduce a technique for computing the fuzzy membership values. If the data affected with noise or outliers then the classification process will influence, so the data needs some preprocessing steps. We propose a method by adding a fuzzy value for those examples that away from the center of the class and the remaining examples don't have any change. Now the new formulation of the problems become as follow:

$$\min_{(w,\gamma)\neq 0} \frac{\|S^A\,Aw - e\gamma\|^2 + \delta \left\|\begin{bmatrix} w \\ \gamma \end{bmatrix}\right\|^2}{\|S^B\,Bw - e\gamma\|^2}, \quad (22)$$

where,

$$G := [S^A\,A \quad -e]'\,[S^A\,A \quad -e] + \delta I, \quad (23)$$

$$H := [S^B B \quad -e]'\,[S^B\,B \quad -e], \qquad z := \begin{bmatrix} w \\ \gamma \end{bmatrix}. \quad (24)$$

The optimization problem (22) becomes:

$$\min_{z\neq 0} r(z) := \frac{z'Gz}{z'Hz}. \quad (25)$$

Similarly we can directly get the second fuzzy plane by solving the following optimization problem.

$$\min_{z\neq 0} s(z) := \frac{z'Lz}{z'Mz}, \quad (26)$$

where,

$$L := [S^B\,B \quad -e]'\,[S^B\,B \quad -e] + \delta I, \quad (27)$$

$$M := [S^A A \quad -e]'\,[S^A\,A \quad -e]. \quad (28)$$

Figure 1 explains the process of computing fuzzy matrix $S^A$.

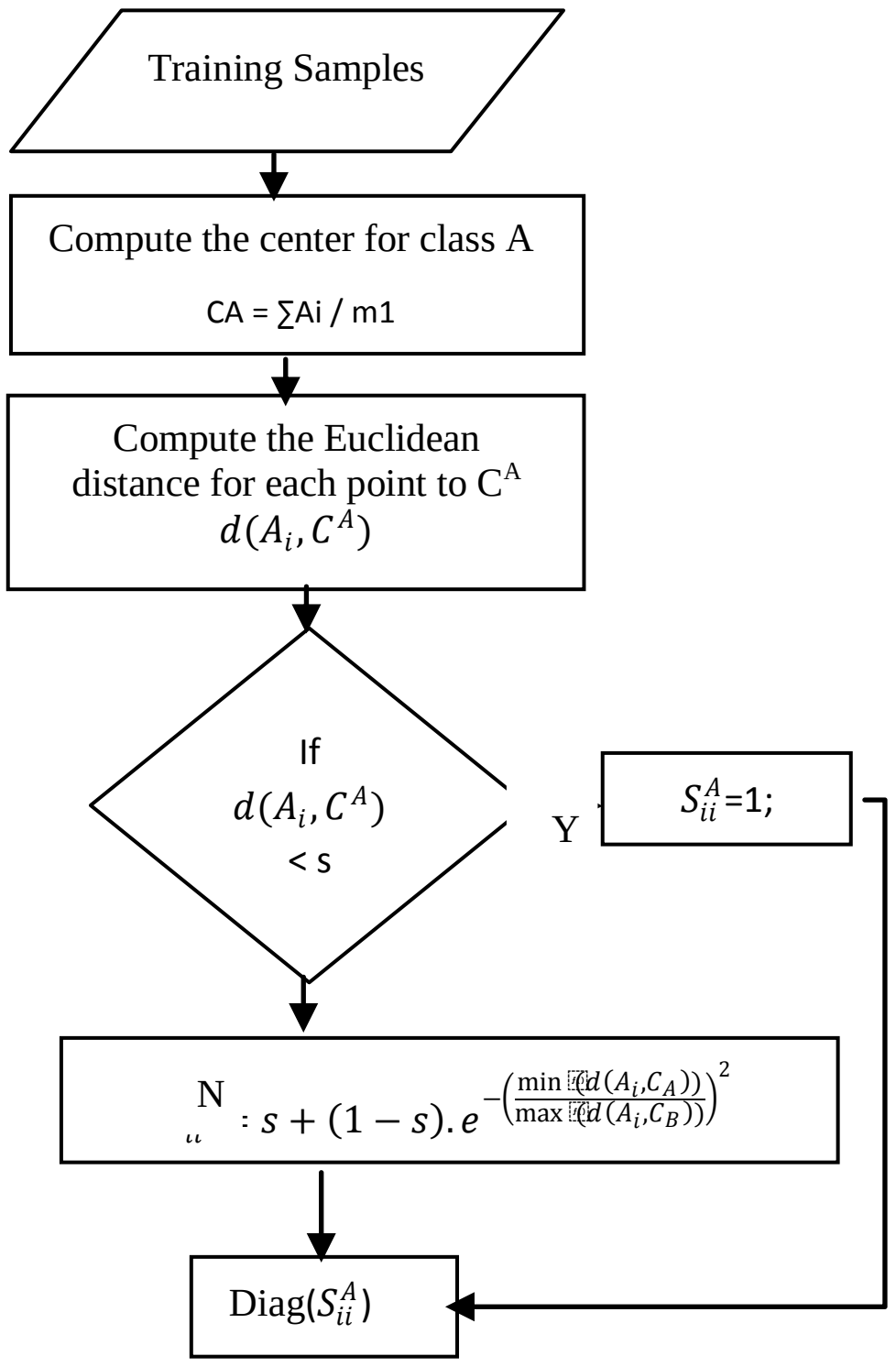

**Fig 1: Methodology of calculation of fuzzy matrix $S^A$**

Where s is a parameter weighting the contribution of the exponential term to $S^A$. The same way for $S^B$. The next step and the most important step in the proposed method is how we can get the optimal parameters. In the linear MSPSVM there is only one parameter $\delta$. DSA algorithm has been used to find the optimal value of $\delta$.

## 4.2 Nonlinear DSA-GEPSVM

Appropriate choice of the kernel function increases the accuracy of the classification. In real life applications the choice of kernel function depends on the dataset used.

Here are some of the most popular kernels.

Polynomial function:

A polynomial kernel is a common method for nonlinear modeling.

$$K(x,x') = (< x,x' > +1)^d. \quad (29)$$

Gaussian radial basis function:

This function has received significant attention, most commonly with a Gaussian of the form,

$$K(x,x') = exp\left(\frac{-\|x-x'\|^2}{2\,\sigma^2}\right). \quad (30)$$

Exponential radial basis function:

$$K(x,x') = exp((-\|x-x'\|\ )/(D\,\sigma\,))\ . \quad (31)$$

In [38] a new kernel has been introduced the author used the new kernel with the standard SVM, in the presented work we use the new kernel which is called PolyRBF witch is a hybrid between a polynomial kernel and a Gaussian RBF kernel.

$$K(x,x') = (1 + \ exp\left(\frac{-\|x-x'\|^2}{D\,\sigma}\right))^d\ . \quad (32)$$

Where D is the dimension of the data, now we have three parameters in the nonlinear case, the first parameter is the regularization of GEPSVM and second and third parameters is for the kernel function if we use the polynomial kernel then we have the parameter d, and if we use the RBF then we have the parameter $\sigma$, last if we use the PolyRBF the we have two parameters d and $\sigma$.

In order to get the nonlinear planes in equation (11), we proposed to use the PolyRBF kernel. In the next subsection we explain how to get the best parameters in the linear and nonlinear classifier.

## 4.3 Parameter Optimization using DSA

A population in DSA assumed to be made up of random solutions of the problem corresponds to an artificial-superorganism migrating. In DSA, artificial- superorganism migrates to the global minimum value of the optimization problem. In the migration time the artificial-superorganism tests whether some positions which was selected randomly are suitable temporary during the migration. Then the process stops over on the suitable tested position for a temporary time during the migration, the members of the artificial that made such discovery immediately settle at the discovered position and continue their migration from this position [28].

In the implementation of DSA, artificial-organisms (i.e., $X_i$, i={1,2,3,…,N}) making up an artificial-superorganism (i.e., Superorganismg, g={1,2,3,…,G}) contain members as much as the size of the problem (i.e., $x_{ij}$, j={1,2,3,…,D}). Where, N signifies number of elements in the superorganism (Size of the population), G represents number of maximum generation, and D indicates size of the problem [28-30].

$$x_{ij} = rand.\left(up_j - \ low_j\right) + \ low_j \quad (33)$$

The stopover site is an important step in migration. The method to find a stopover site at the remaining between the artificial- organisms may be described by a Brownian-like random walk model [28]. By a random selection of individuals of the artificial- organisms move toward the targets of $donor = \left[X_{random\ _shuffling\ (i)}\right]$ to discover stopover sites. The scale value (R) is used to control the size of the change occurred in the positions of members of the artificial-organisms. The way of calculation R makes the respective artificial-superorganism to radically change direction in the habitat [28-30].

The stopover site position in DSA is produced by using equation (34):

$$StopoverSite = Superorganism + Scale * (donor - Superorganism) \quad (34)$$

A random process is used to determine the members of the artificial organisms of the superorganism of stopover site. If the one of the stopover site elements goes outside the limits of the search space for any reason, it randomly deferred to another position in the search space. If the stopover site is better than the sources owned by the artificial-organism, the artificial-organism moves to that stopover site. While the artificial-organisms change site, the superorganism containing the artificial organisms continues its migration to the global minimum.

There are two control parameters in DSA, which are p1 and p2. The tested and the most appropriate values for these parameters were conducted by [28]. Figure 2 describes the main steps of the proposed approach.

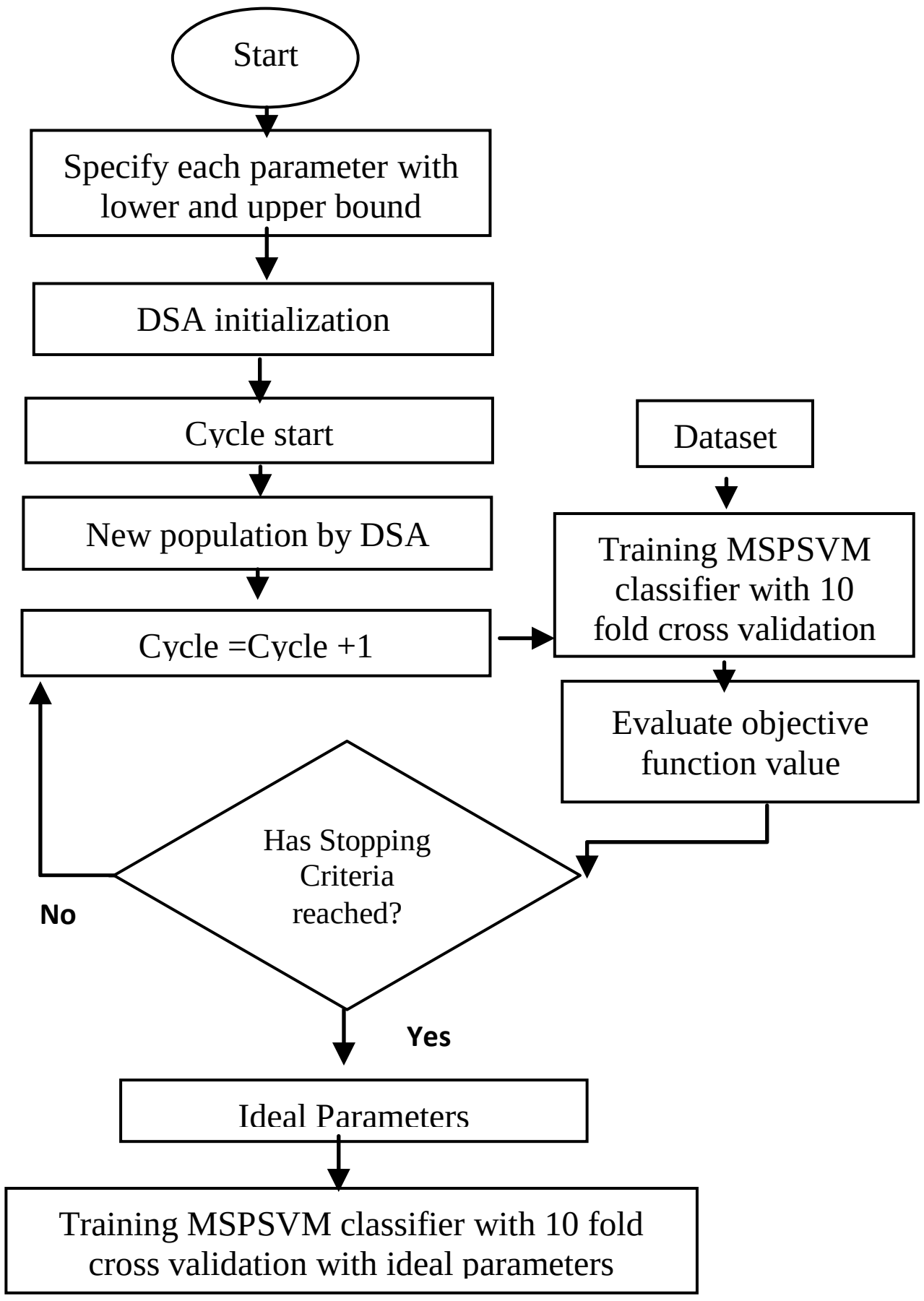

**Fig 2: Basic steps of the proposed approach**

In the proposed approach we are seeking for optimal values for the linear and nonlinear parameters. The stopping criteria for the procedure is either max number of cycles are reached or the 100% accuracy is obtained.

# 5. EXPERIMENTAL RESULTS

The proposed approach is implemented on personal computer with a core i3 processor 2.13GHz, 3GB of RAM, and windows 8.1 operating system. Matlab 2010b framework is used in development. To verify the proposed approach quality the following datasets are used from UCI repository [39], the datasets characteristics are shown in Table 1.

**Table 1 Datasets characteristics.**

| Dataset | Features | Instances |
|---|---|---|
| Australian | 14 | 690 |
| Breast Cancer | 10 | 683 |
| Diabetes | 8 | 768 |
| German | 24 | 1000 |
| Heart Disease | 13 | 270 |
| Ionosphere | 34 | 351 |
| Liver Disorders | 6 | 345 |
| Sonar | 60 | 208 |
| Splice | 60 | 1000 |
| Wbpc | 34 | 198 |

Tables 2 and 6 summarize all parameters setting in the linear and nonlinear DSA-MSPSVM respectively with their assigned values. Where the values are chosen based on our numerical experiments.

For implementation, the data was divided into ten parts or folds, nine of which comprised the training data, with the tenth being used for testing the generalization ability of the classifier. This process was repeated ten times, using a different fold for testing on each occasion. This process is known as tenfold cross validation and is a standard methodology for reporting the performance of a classifier. The classification accuracy was computed by computing the average across all the ten trials.

**Table 2 Linear GEPSVM parameter**

| Parameter | Symbol | Interval |
|---|---|---|
| P1 | $\delta$ | [0.001,10000] |

The DSA parameters setting are shown in table 3.

**Table 3 DSA-GEPSVM parameters setting**

| DSA Parameter | Definition | Value |
|---|---|---|
| Popsize | Size of superorganism | 30 |
| Dim | Dimension of search space | 1 |
| Low | Minimum limit of search space | 0.001 |
| Up | Maximum limit of search space | 10000 |
| Maxcycle | Max number of Iteration | 20 |

Table 4 illustrates the results obtained after implementing Linear DSA-GEPSVM on several public benchmark datasets.

**Table 4 Linear DSA-GEPSVM accuracy results on some benchmark datasets**

| Dataset | Proposed method | | |
|---|---|---|---|
| | P1 | Training | Testing |
| Australian | 920.9123 | 70.6924 | 73.913 |
| Breast Cancer | 4.8162 | 97.3984 | 100 |
| Liver Disorders | 964.9806 | 70.7395 | 77.1429 |
| Diabetes | 5 | 74.2775 | 80.5195 |
| German | 15.8707 | 74.7778 | 82 |
| Heart Disease | 0.8947 | 87.2428 | 96.2963 |
| Ionosphere | 0.1306 | 81.132 | 76.064 |
| Sonar | 0.2105 | 91.9786 | 90.4762 |
| Splice | 40.3143 | 74.8889 | 76 |
| wpbc | 0.001 | 89.3258 | 95 |

Table 5 illustrates the comparison between Linear DSA-GEPSVM and four recently methods GEPSVM, FSVM, FTSVM, and IGEPSVM. The proposed method given promising results for all dataset from other methods, and the mean accuracy of proposed method is the best.

**Table 5 Training accuracy of linear DSA-GEPSVM and compared methods on UCI datasets**

| Dataset | Proposed | GEPSVM [12] | FSVM [13] | FTSVM [18] | IGEPSVM [36] |
|---|---|---|---|---|---|
| Australian | 70.6924 | - | 85.56 | **86.08** | - |
| Breast Cancer | **97.3984** | - | 65. 01 | 65.60 | - |
| Liver Disorders | 70.7395 | 68.86 | 76.67 | **77.80** | 73.83 |
| Diabetes | 74.2775 | 67.93 | - | - | **74.61** |
| German | 74.7778 | 75.49 | 71.68 | **78.20** | 77.15 |
| Heart Disease | **87.2428** | - | 83.33 | 84.44 | - |
| Sonar | **91.9786** | 83.66 | - | - | 88.47 |
| wpbc | **89.3258** | 83.98 | - | - | 87.74 |
| mean | **82.0541** | 75.984 | 79.31 | 78.424 | 80.36 |

The deferent kernels were applied to nonlinear DSA-GEPSVM. The first kernel is the polynomial kernel, the second kernel is the radial base function kernel and the last one the hybrid kernel between the previously mentioned kernels. In order to prove how the hybrid kernel is effective, we applied the nonlinear DSA-GEPSVM three times on each kernel. Table 7 shows the detailed results that obtained. Figure 3 shows how the hybrid kernel is effective in most cases.

**Table 6: Nonlinear DSA-GEPSVM parameters**

| Parameter | Symbol | Interval |
|---|---|---|
| P1 | $\delta$ | [0.001,10000] |
| P2 | $\sigma$ | [0.001,33] |
| P3 | p | [0.001,33] |

**Table 7: The nonlinear DSA-GEPSVM results applied to the three mentioned kernels**

| Dataset | Kernel | P1 | P2 | P3 | ACC Training | ACC Testing | Mean ACC |
|---|---|---|---|---|---|---|---|
| Breast Cancer | rbf | 562.9775 | 0.001 | - | 100.00 | 98.55 | 99.28 |
| | poly | 1000.0000 | - | 4.5906 | 100.00 | 100.00 | 100.00 |
| | PolyRBF | 125.4612 | 24.8243 | 5.3731 | 96.92 | 98.55 | 97.73 |
| Heart | rbf | 0.0010 | 4.5501 | - | 87.24 | 88.89 | 88.07 |
| | poly | 1000.0000 | - | 8.9368 | 100.00 | 81.48 | 90.74 |
| | PolyRBF | 2516.2190 | 12.9724 | 25.134 | 100.00 | 85.19 | 92.59 |
| Ionosphere | rbf | 20.6282 | 0.001 | - | 100.00 | 94.29 | 97.14 |
| | Poly | 0.0010 | - | 4.1069 | 100.00 | 80.00 | 90.00 |
| | PolyRBF | 841.7760 | 32.5201 | 24.1597 | 100.00 | 97.06 | 98.53 |
| Sonar | rbf | 13.1398 | 0.001 | - | 100.00 | 90.00 | 95.00 |
| | poly | 8884.2091 | - | 6.3464 | 100.00 | 76.19 | 88.10 |
| | PolyRBF | 0.0010 | 19.9701 | 19.631 | 100.00 | 85.71 | 92.86 |

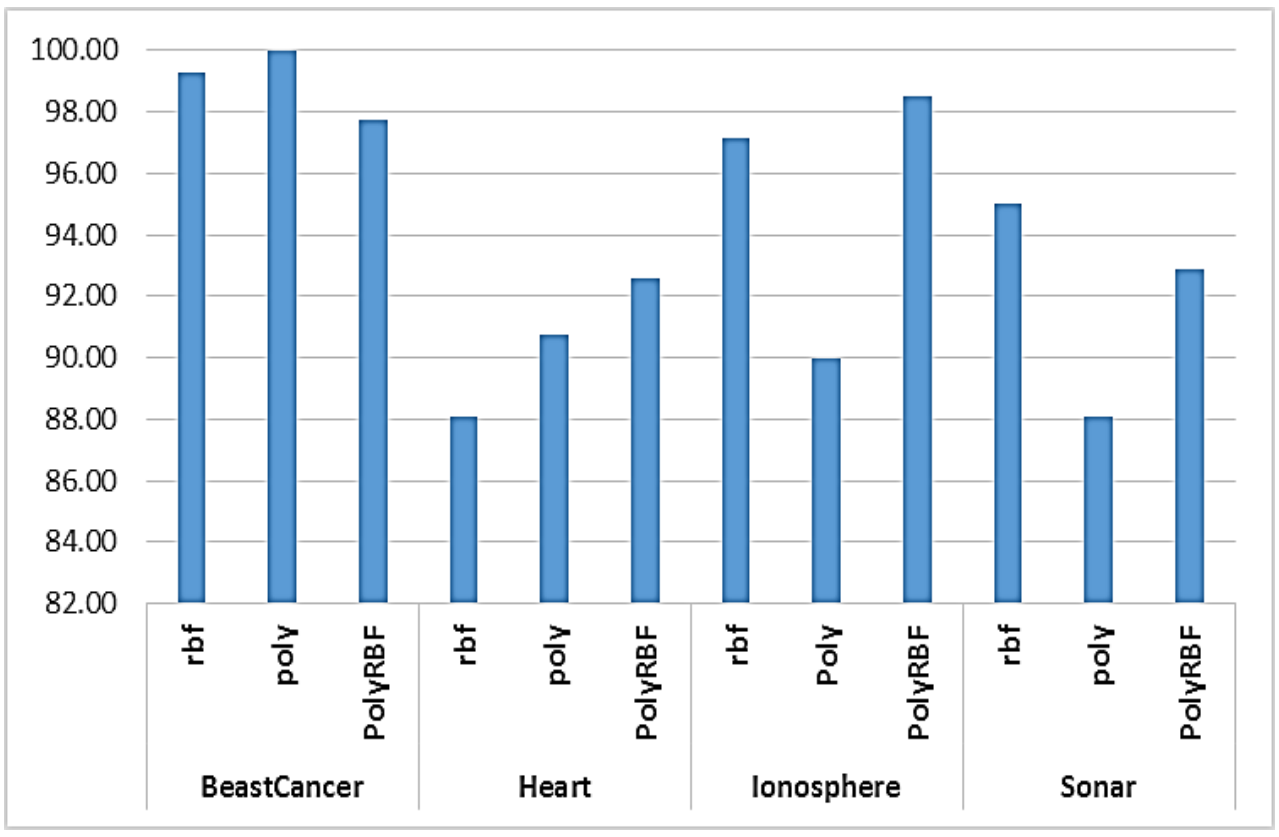

**Fig 3: Mean accuracies of poly, rbf, and PolyRBF applied on UCI datasets**

Table 8 and Table 9 illustrate the comparison between training and testing accuracies of nonlinear DSA-GEPSVM on four datasets with other recently methods GA+SVM, SA+SVM, PSO+SVM, CV-ACC, and S.C. Chen. Results proved how the proposed approach give comparable and promising results.

**Table 8: Training accuracies comparisons between the nonlinear DSA-GEPSVM with PolyRBF and other recently approaches**

| Method | Datasets | | | |
|---|---|---|---|---|
| | Breast Cancer | Heart Disease | Ionosphere | Sonar |
| Proposed | 96.92 | **100** | **100** | **100** |
| GA+SVM[20] | 94.23 | 94.58 | 96.61 | 95.22 |
| SA+SVM[21] | **97.95** | 87.97 | 97.5 | 91.85 |
| PSO+SVM[22] | **97.95** | 88.17 | 97.5 | 88.32 |
| CV-ACC[23] | 96.69 | 84.753 | 97.714 | **100** |
| S.C. Chen [24] | 96.04 | 86.32 | 96.60 | 96.07 |

**Table 9: Testing accuracies comparisons between the nonlinear DSA-GEPSVM with PolyRBF and other recently approaches**

| Method | Datasets | | | |
|---|---|---|---|---|
| | Breast Cancer | Heart Disease | Ionosphere | Sonar |
| Proposed | **98.55** | 85.19 | 97.06 | 85.71 |
| GA+SVM[20] | 94.23 | **94.58** | 96.61 | 95.22 |
| SA+SVM[21] | 97.95 | 87.97 | **97.5** | 91.85 |
| PSO+SVM[22] | 97.95 | 88.17 | **97.5** | 88.32 |
| CV-ACC[23] | 95.97 | 83.98 | 93.68 | 87.26 |
| S.C. Chen [24] | 96.96 | 91.15 | **97.5** | **96.90** |

## 6. CONCLUSIONS

In this paper, we presented a DSA-MSPSVM method for data classification based on MSPSVM and DSA approaches. It is well known that the MSPSVM regularization parameter $\delta$ and kernel parameters are important to the performance of the classifier. But it is difficult to choose a kernel function and its parameters because they are dependent on datasets. The DSA has been applied to optimize these parameters. We conducted experiments to evaluate the performance of the proposed approach with three different kernel functions Poly, RBF, and PolyRBF in the nonlinear classifier. The results obtained were compared with those obtained with other algorithms. The results show enough evidence that the proposed approach has less error rates across most of the datasets with other algorithms. We can also conclude that PolyRBF kernel gives better results as compared with other kernel functions. Further, we plan to extend the DSA-MSPSVM approach to deal with multiclass problems in the linear and nonlinear cases, and study the kernel function effects in the datasets.

## 8. APPENDIX

**Pseudo code: Differential search algorithm [28]**

Require:

N: Size of the population, where i = {1, 2, 3, …, N}

D: Dimension of the problem

G: Number of maximum generation

```
Superorganism = initialize(), where Superorganism = [ArtificialOrganism_i]
y_i = Evaluate(ArtificialOrganism_i )
for cycle = 1: G do
    donor = Superorganism_Random_Shuffling(i)
    Scale   = randg[2.rand_1] . (rand_2 -  rand_3)
    StopoverSite =Superorganism +Scale . (donor - Superorganism)
    p_1=0.3 . rand_4 and  p_2=0.3 . rand_5
    if  rand_6 < rand_7 then
        if  rand_8 < p_1 then
            r = rand(N,D)
            for Counter1=1 : N do
                r(Counter1,:) = r(Counter1,:) < rand_9
            end for
        else
            r = ones(N,D)
            for Counter2=1 : N do
                r(Counter2, randi(D)) = r(Counter2, randi(D)) <rand_10
            end for
        end if
    else
        r = ones(N,D)
        for Counter3=1 : N do
            d = randi(D,1,⌈p2 . rand . D⌉)
            for Counter4 = 1 : size(d) do
                r(Counter3,d(Counter4)) = 0
            end for
        end for
```

**end if**

$individuals_{I,J} \leftarrow r_{I,J} > 0 \mid I \in i, J \in [1\ D]$

$StopoverSite(individuals_{I,J}) := Superorganism(individuals_{I,J})$

**if** $StopoverSite_{i,j} < low_{i,j}$ **or** $StopoverSite_{i,j} > up_{i,j}$ **then**

$StopoverSite_{i,j} = rand \cdot (up_j - low_j) + low_j$

**end if**

$y_{StopoverSite;i} = evaluate(StopoverSite_i)$

$$y_{Superorganism;i} = \begin{cases} y_{StopoverSite;i} & if\ y_{StopoverSite;i} < y_{Superorganism;i} \\ y_{Superorganism;i} & else \end{cases}$$

$$ArtificialOrganism_i = \begin{cases} StopoverSite_i & \text{if } y_{StopoverSite;i} < y_{Superorganism;i} \\ ArtificialOrganism_i & else \end{cases}$$

**end for**